\documentclass[11pt,a4paper,twocolumn]{article}
\usepackage{times}
\usepackage{latexsym}
\usepackage{here}
\usepackage{amsmath}
\usepackage{amsfonts}
\usepackage{amsthm}
\usepackage{amssymb}
\usepackage{color}
\usepackage[utf8]{inputenc}
\usepackage[english]{babel}
\usepackage[linesnumbered,ruled,vlined]{algorithm2e}
\usepackage{multirow}
\usepackage{tikz}
\usepackage{booktabs}
\usepackage{microtype}
\usepackage[hyphens]{url}

\newtheorem{defi}{Definition}

\newcommand{\argmin}{\mathop{\mathrm{argmin\,}}}

\newcommand{\mathbbR}{\mathbb{R}}

\newcommand{\boldone}{{\boldsymbol{1}}}

\newcommand{\boldD}{{\boldsymbol{D}}}

\newcommand{\boldP}{{\boldsymbol{P}}}

\newcommand{\bolda}{{\boldsymbol{a}}}
\newcommand{\boldb}{{\boldsymbol{b}}}

\newcommand{\boldh}{{\boldsymbol{h}}}

\newcommand{\boldp}{{\boldsymbol{p}}}

\newcommand{\boldw}{{\boldsymbol{w}}}
\newcommand{\boldx}{{\boldsymbol{x}}}
\newcommand{\boldy}{{\boldsymbol{y}}}

\newcommand{\boldtheta}{{\boldsymbol{\theta}}}

\newcommand{\boldPi}{{\boldsymbol{\Pi}}}

\newcommand{\calL}{{\mathcal{L}}}

\newcommand{\calP}{{\mathcal{P}}}

\newcommand{\calT}{{\mathcal{T}}}

\newcommand{\calX}{{\mathcal{X}}}
\newcommand{\calY}{{\mathcal{Y}}}

\title{Computationally Efficient Wasserstein Loss for Structured Labels}

\author{
Ayato Toyokuni${}^{\,1,3}$
\quad\hspace{-5pt}
Sho Yokoi${}^{\,2,3}$
\quad\hspace{-5pt}
Hisashi Kashima${}^{\,1,3}$
\quad\hspace{-5pt}
Makoto Yamada${}^{\,1,3}$
\\
${}^{1}$ Kyoto University \quad
${}^{2}$ Tohoku University \quad
${}^{3}$ RIKEN AIP\\
{\small \tt \{toyokuni.ayato@ml.ist.i,\hspace{0.2em}kashima@i,\hspace{0.2em}myamada@i\}.kyoto-u.ac.jp,}\\
{\small \tt yokoi@ecei.tohoku.ac.jp}
}

\date{}

\begin{document}
\maketitle
\begin{abstract}
 The problem of estimating the probability distribution of labels has been widely studied as a label distribution learning (LDL) problem, whose applications include age estimation, emotion analysis, and semantic segmentation. We propose a tree-Wasserstein distance regularized LDL algorithm, focusing on hierarchical text classification tasks. We propose predicting the entire label hierarchy using neural networks, where the similarity between predicted and true labels is measured using the tree-Wasserstein distance.
 Through experiments using synthetic and real-world datasets, we demonstrate that the proposed method successfully considers the structure of labels during training, and it compares favorably with the Sinkhorn algorithm in terms of computation time and memory usage.
 
\end{abstract}

\section{Introduction}

Label distribution learning (LDL), which is a generalized framework for performing single/multi-label classification and estimating the probability distribution over labels, is an important machine-learning problem \cite{geng2016label}. Its applications include age estimation \cite{geng2013facial}, emotion estimation \cite{DBLP:conf/emnlp/ZhouZZZG16}, head-pose estimation \cite{geng2014head}, and semantic segmentation \cite{gao2017deep}. In particular, multi-label classification is an important problem in many NLP areas, and has several applications including multi-label text classification \cite{DBLP:conf/acl/BanerjeeAPT19,DBLP:conf/acl/ChalkidisFMA19}.

Typically, Kullback-Leibler (KL) divergence is used to measure the similarity between two distributions. However, the KL divergence can tend to infinity if the supports of the two distributions do not overlap, resulting in model failure.

To solve this support problem, Wasserstein distance is used instead of KL divergence \cite{pmlr-v70-arjovsky17a}. 
Wasserstein distance is defined as the cost of optimally transporting one probability distribution to match another \cite{villani2009optimal,peyre2018computational}. Because it can compare two probability measures while considering the ground metric, it is more powerful than measurements that do not consider geometrical information.

An LDL framework with Wasserstein distance has been recently proposed \cite{frogner2015learning,DBLP:conf/aaai/ZhaoZ18a}. 
This framework employs the Sinkhorn algorithm \cite{cuturi2013sinkhorn} to calculate the Wasserstein distance, which requires quadratic computational-time. Thus, when we consider extremely large label-sets, for example, $10^5$, the computation cost can be significant. However, the Wasserstein distance on a tree (hereinafter called {\it tree-Wasserstein distance}) can be written in a closed-form and calculated in linear computation time \cite{evans2012phylogenetic,le2019tree}.

In this paper, we propose a tree-regularized LDL algorithm with a tree-Wasserstein distance. The key advantage of the tree-Wasserstein distance is that it considers the hierarchical label information explicitly, whereas the Sinkhorn-based algorithm needs a cost matrix using tree-structured data. Moreover, the tree-Wasserstein distance has an analytic form that can be computed in linear time using significantly less memory. We experimentally demonstrate that the proposed algorithm compares favorably with the Sinkhorn-based LDL algorithm \cite{frogner2015learning,DBLP:conf/aaai/ZhaoZ18a} with considerably lower memory consumption and computational costs. We demonstrate that the calculation is more efficient than that of the existing Wasserstein loss.

\vspace{.1in}
\noindent {\bf Contribution:} Our contributions are summarized as follows. (1) We propose training a model by minimizing the tree-Wasserstein distance for hierarchical labels, and (2) we experimentally show that the proposed method is computationally more efficient than the existing methods with Sinkhorn-based methods.

\section{Problem Setting}\label{sec-problem}
We observe $n$ input and output samples $\{(\boldx_1, \boldy_1),\cdots,(\boldx_n,\boldy_n)\}$ from $(\calX,\calY)$, where $\mathcal{X} \subset \mathbb{R}^d$.  We consider the problem of learning a map from a feature space  $\calX$ into $\calP$, which is a set of distributions over a finite set $\mathcal{Y}$.

For example, multi-class classification is included in this problem, $\boldy$, which represents the $\ell$-th class, and it is expressed as the following one-hot vector:
\[
\boldy = (0,\ldots,0, \underbrace{1}_{\ell}, 0,\ldots 0)^\top \in \mathbbR^L,
\]
where $L$ denotes the total number of classes, and $\boldy^\top \boldone_L = 1$. Additionally, $\boldone_L \in \mathbbR^L$ denotes a vector whose elements are all $1$.

When multi-label classification is considered, $\calP$ denotes binary vectors that indicate existing labels. For example, if the sample $\boldx$ belongs to classes $\ell$ and $\ell'$, $\boldy$ is given as
\footnotesize
\[
\boldy = (0,\ldots,0, \underbrace{1}_{\ell}, 0,\ldots 0, \underbrace{1}_{\ell'}, 0, \ldots, 0)^\top \in \mathbbR^L,
\]
\normalsize
where $\boldy^\top \boldone_L = 2$. Accordingly, we can transform $\boldy$ into a probability vector as $\boldp_\boldy = \boldy/\boldy^\top \boldone_L$. Notably, we assume that $\mathcal{Y}$ is discrete and has a tree structure similar to hierarchical labels. 

We aim to estimate the conditional probability vector $\boldp_\boldy$ for $\boldx$ by considering the structure information of $\mathcal{Y}$ from $\{(\boldx_1, \boldp_{\boldy_1}),\cdots,(\boldx_n,\boldp_{\boldy_n})\}$.

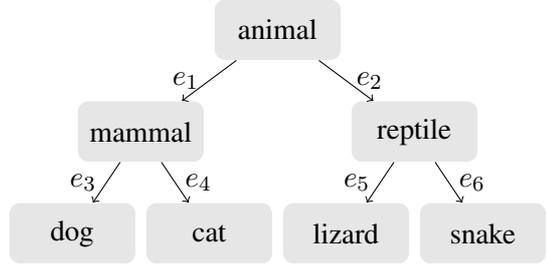
\begin{figure}[t]
    \centering
    \begin{tikzpicture}[scale=0.9]
    \tikzset{block/.style={rectangle, fill=black!10, text centered, rounded corners, text width=1.4cm, minimum height=0.8cm}};
    \node[block] {animal} [level distance=1.5cm, sibling distance=4cm]
        child{ node[block] {mammal}[level distance=1.5cm, sibling distance=2cm]
            child {node[block] {dog} edge from parent [->] node[left] {$e_3$}}
            child {node[block] {cat} edge from parent [->] node[right] {$e_4$}}
            edge from parent [->] node[left] {$e_1$}
        }
        child{ node[block] {reptile}[level distance=1.5cm, sibling distance=2cm, edge from parent/.style={<-,draw}]
            child {node[block] {lizard} edge from parent [->] node[left] {$e_5$}}
            child {node[block] {snake} edge from parent [->] node[right] {$e_6$}}
            edge from parent [->] node[right] {$e_2$}
        };
 \end{tikzpicture}
    \caption{Illustration of a tree-structured label with the root ``animal''. $\Gamma({\rm "mammal"}) = \{{\rm ``mammal"}, {\rm ``dog"}, {\rm ``cat"}\}$,\ $v_{e_2} = {\rm ``reptile"}$\ .}
    \label{fig:tree}
\end{figure}

\begin{table*}[ht]
	\begin{center}
		    \caption{The results for the Synthetic dataset. The label distributions are given on a random tree with $1000$ nodes. \label{tb:syntree}}
		\scalebox{0.65}{
		\begin{tabular}{cccccccccc} \toprule
			\multicolumn{1}{c}{Loss} & Wasserstein $\downarrow$ & KL $\downarrow$ & Cheby$\downarrow$ & Clark $\downarrow$ & Canbe $\downarrow$ & Cos $\uparrow$ & IntSec $\uparrow$\\ \midrule
			\multicolumn{1}{c}{$\mathcal{KL}$} & $9.701 \pm(.050)$ & $\bf{0.431 \pm(.001)}$ & $0.209 \pm(.001)$ & $1.777 \pm(.011)$ & $14.512 \pm(.060)$ & $\bf{0.877 \pm(.000)}$ & $\bf{0.754 \pm(.001)}$\\
			
			 $\mathcal{KL} + \frac{1}{2}\mathcal{W}^1$ & $10.831 \pm(.044)$ & $0.452 \pm(.001)$ & $0.230 \pm(.001)$ & $1.666 \pm(.009)$ & $13.834 \pm(.064)$ & $0.868 \pm(.000)$ & $0.739 \pm(.001)$\\
	         $\mathcal{KL} + \mathcal{W}^1$ & $11.631 \pm(.048)$ & $0.475 \pm(.001)$ & $0.244 \pm(.001)$ & $\bf{1.618 \pm(.008)}$ & $\bf{13.474 \pm(.063)}$ & $0.859 \pm(.000)$ & $0.727 \pm(.001)$\\
			
			 $\mathcal{KL} + \frac{1}{2}\mathcal{TW}$ & $7.257 \pm(.110)$ & $0.595 \pm(.007)$ & $\bf{0.193 \pm(.001)}$ & $2.098 \pm(.040)$ & $19.636 \pm(.171)$ & $0.833 \pm(.002)$ & $0.729 \pm(.003)$\\
			 $\mathcal{KL} + \mathcal{TW}$ & $\bf{7.158 \pm(.117)}$ & $0.631 \pm(.007)$ & $0.195 \pm(.001)$ & $2.143 \pm(.030)$ & $19.923 \pm(.441)$ & $0.825 \pm(.003)$ & $0.721 \pm(.004)$ \\
			 \bottomrule
		\end{tabular}
	}
		\hspace{10cm}
		\vspace{-.1in}
	\end{center}
\end{table*}
\begin{table}[ht]
\small
	\begin{center}
		\caption{The results for BlurbGenreCollectionEN. \label{tb:blurb}}
	\scalebox{0.8}{
		\begin{tabular}{cccccc} \toprule
			Loss & Pseudo-Recall & Top5 & AUC\\ \midrule
			$\mathcal{KL}$ & $0.679 \pm(.008)$ & $1.013 \pm(.015)$ & $0.971 \pm(.001)$\\ 
			 $\mathcal{KL} + \frac{1}{2}\mathcal{W}^1$ & $0.675 \pm(.008)$ & $1.009 \pm(.013)$ & $0.970 \pm(.002)$\\
			 $\mathcal{KL} + \mathcal{W}^1$  & $0.678 \pm(.004)$ & $1.008 \pm(.018)$ & $0.970 \pm(.001)$ \\
		     $\mathcal{KL} + \frac{1}{2}\mathcal{TW}$ & $0.678 \pm(.010)$ & $0.993 \pm(.013)$ & ${0.971} \pm(.002)$\\
		     $\mathcal{KL} + \mathcal{TW}$ & $0.678 \pm(.009)$ & ${0.991} \pm(.017)$ & $0.970 \pm(.001)$\\
			 \bottomrule

		\end{tabular}
	}
	\end{center}
	\vspace{-.1in}
\end{table}

\section{Proposed Method}\label{sec-proposed}
In this study, we assume $\calY$ has a tree-structure. Accordingly, we propose LDL with tree-Wasserstein distance. 
\subsection{Wasserstein distance on tree metrics}\label{subsec-wdtree}

Let $\mathcal{T}$ be a tree with non-negative weighted edges and $\mathcal{N}_\mathcal{T}$ be the set of nodes of $\mathcal{T}$. A shortest path metric $d_{\mathcal{T}} : \mathcal{N}_\mathcal{T} \times \mathcal{N}_\mathcal{T} \to \mathbb{R}$ associated with $\mathcal{T}$ is called the {\it tree metric}. Let $v$ and $v'$ be the nodes in $\calT$. Accordingly, $d_\mathcal{T}(v, v')$ is equal to the sum of the edge weights along the shortest path between $v$ and $v'$. Next, we know that $\mathcal{M}_\mathcal{T} = (\mathcal{N}_\mathcal{T}, d_\mathcal{T})$ is a metric space and can be naturally derived from $\mathcal{T}$.

It is assumed that $\mathcal{T}$ is rooted at $r$. For each node $v$, the set of nodes in the sub-tree of $\mathcal{T}$ rooted at $v$ is defined as $\Gamma(v) = \{u \in \mathcal{N}_\mathcal{T} \mid v \in \mathcal{R}(u)\}$  where $\mathcal{R}(v)$ denotes the set of nodes in a unique path from a node $v$ to the root $r$ in $\mathcal{T}$. For each edge $e$, $v_e$ denotes a deeper level node. Figure \ref{fig:tree} illustrates a tree-structured label.

Given two probability measures $\mu, \nu$ supported on $\mathcal{M}_\mathcal{T}$, the 1-Wasserstein distance between $\mu$ and $\nu$ is expressed as follows \cite{evans2012phylogenetic,le2019tree}: 
\begin{equation}
    \mathcal{W}_{d_{\mathcal{T}}}^1(\mu, \nu) = \sum_{e \in \mathcal{T}} w_e |\mu(\Gamma(v_e)) - \nu(\Gamma(v_e))|,
\end{equation}
where $w_e$ denotes the weight of edge $e$. The key advantage of the tree-Wasserstein distance is that it can be computed with the linear time complexity, whereas the time complexity for the Sinkhorn algorithm is quadratic \cite{cuturi2013sinkhorn}.

\subsection{LDL with tree-Wasserstein distance}

We define the tree-Wasserstein regularizer as follows.

\begin{defi}[tree-Wasserstein regularizer]
Let $\boldh_\boldtheta : \mathcal{X} \to \calP$ be a model with learnable parameters $\theta$. Let $T_\mathcal{Y} = (V, E, W_E)$ be a tree associated with $\mathcal{Y}$, where $V$ denotes the set of nodes, $E$ is the set of edges, and $W_E(e)$ is the length of edge $e \in E$. Given input $\boldx \in \mathcal{X}$ and the ground-truth distribution of $\boldy$ $\boldp_\boldy \in \calP$, then the tree-Wasserstein regularization term $\mathcal{TW}(\boldx, \boldp_\boldy)$ is defined as follows:
\begin{align*}
        &\mathcal{TW}(\boldx, \boldp_\boldy) \nonumber\\
        &\!\!= \sum_{e \in \mathcal{T}} W_E(e) |(\boldh_\boldtheta(\boldx))(\Gamma(v_e)) 
        \!-\! \boldp_\boldy(\Gamma(v_e))|,
\end{align*}
where $\boldh_\boldtheta$ denotes the prediction model.
\end{defi}

Using the tree-Wasserstein regularizer, we propose the following LDL:
\begin{align}
    \label{eq:treereg}
    \widehat{\boldtheta} = \argmin_{\boldtheta} &\sum_{i = 1}^{n} \lambda \mathcal{TW}(\boldh_\boldtheta(\boldx_i), \boldp_{\boldy_i}) \nonumber \\
    &+ \mathcal{KL}(\boldh_\boldtheta(\boldx_i),\boldp_{\boldy_i}),
\end{align}
where 
\begin{align}
\label{eq:kl}
\mathcal{KL}(\boldh_\boldtheta(\boldx_i),\boldp_{\boldy_i}) = \sum_{\ell = 1}^L \boldp_{\boldy_i}^{(\ell)} \log \frac{\boldp_{\boldy_i}^{(\ell)}}{\boldh_\boldtheta(\boldx_i)^{(\ell)}},
\end{align}
is the multi-class Kullback-Leibler loss function, and $\lambda \geq 0$ is its regularization parameter. 

Notably, $\mathcal{TW}(\boldh_\boldtheta(\boldx_i), \boldp_{\boldy_i})$ is calculated in $O(L)$ time, where $L$ denotes the number of labels. Unlike the Sinkhorn-Knopp algorithm, we need not compute and hold a distance matrix. For tree-structured labels, including hierarchical labels, the tree structure can be used directly as a tree metric. If we have prior knowledge about labels (e.g., similarity), we can set edge-weights using the prior knowledge. 

\section{Related Work}\label{sec-related} 

\subsection{Label distribution learning}

LDL \cite{geng2016label} is the task of estimating the distribution of labels from each input. While age estimation \cite{geng2013facial}, head-pose estimation \cite{geng2014head}, and semantic segmentation \cite{gao2017deep} are well known LDL tasks, in this study, we consider the task of estimating a distribution on a hierarchical structure. The key difference between LDL and a generative model is that the ``true'' distribution on labels is given in LDL.

\subsection{Wasserstein distance}

Given two probability vectors $\bolda, \boldb \in \mathbb{R}_{\geq 0}^n$ and a distance matrix $\boldD\in \mathbb{R}^{n\times n}_{\geq 0}$, the 1-Wasserstein distance $\mathcal{W}^1(\bolda, \boldb)$ between $\bolda$ and $\boldb$ is defined as:
\begin{align}
    \mathcal{W}^1(\bolda, \boldb) = \min_{\boldP \in \Pi}\langle \boldD, \boldP \rangle, \label{eq:dwassdis}
\end{align}
where $\boldPi$ denotes the set of transport plans such that $\boldPi = \{\boldP \in \mathbb{R}_{\geq 0}^{n \times n} \mid \boldP \boldone_n = \bolda, \boldP^\top \boldone_n = \boldb\}$.

Because Wasserstein distance can incorporates the ground metric in the comparison of the probability distributions, it has been widely used in applications, including domain adaptation \cite{courty2016optimal}, generative models \cite{pmlr-v70-arjovsky17a}, and natural language processing \cite{kusner2015word}.
A loss function that uses the Wasserstein distance can improve predictions based on a structure of labels \cite{frogner2015learning, DBLP:conf/aaai/ZhaoZ18a}. Additionally, an entropic optimal transport loss can provide a robustness against noise labels by finding the coupling of the data samples and propagating their labels according to the coupling weight \cite{damodaran2019entropic}.

Frogner et al. \cite{frogner2015learning} proposed learning using a Wasserstein loss to consider the geometric information in predicting a probability distribution. Because computing a sub-gradient of the exact Wasserstein loss is expensive, they estimated the sub-gradient by introducing an entropic-regularization term and using the Sinkhorn-Knopp algorithm. Although they also suggested extending the Wasserstein loss to unnormalized measures, we do not consider this case. Zhao and Zhou \cite{DBLP:conf/aaai/ZhaoZ18a} showed that Wasserstein loss influenced LDL in terms of simultaneously learning label correlations and distribution.
We proposed learning using an exact Wasserstein distance with efficient computations when the ground metric is represented by a tree.

Le et al. \cite{le2019tree} suggested the tree-sliced Wasserstein distance, where the Wasserstein distance is approximated on a continuous space by averaging the Wasserstein distances on tree metrics constructed by dividing that space. An unbalanced variant of the tree-Wasserstein distance has been recently proposed \cite{sato2020fast}.

\section{Experiments}\label{sec-experiment}
We applied our proposed method to LDL on trees based on a synthetic dataset and to multi-label text classification of a hierarchical structure based on a real dataset. We implemented all the methods using Pytorch \cite{NEURIPS2019_9015}. Our models were optimized using a gradient method with the {\it Adam} \cite{DBLP:journals/corr/KingmaB14} optimizer.

\vspace{.05in}
\noindent {\bf Baselines:}
  We compared our proposed method to the Wasserstein-loss-based LDL framework \cite{frogner2015learning,DBLP:conf/aaai/ZhaoZ18a} and a multi-class KL loss mentioned in (\ref{eq:kl}). 
Notably, in the original paper \cite{DBLP:conf/aaai/ZhaoZ18a}, they did not include KL loss and used only Wasserstein loss, but \cite{frogner2015learning} used a linear combination of KL divergence and Wasserstein distance as the loss.
  To ensure fair comparison, we also report the combination of Wasserstein loss and multi-class KL loss as a strong baseline. Therefore, we set the combination parameter $\lambda = \{0, \frac{1}{2}, 1\}$ defined in Eq \ref{eq:treereg} and the weight of all edges to $1$. The Wasserstein loss was computed using the Sinkhorn-Knopp algorithm in the log domain\cite{schmitzer2019stabilized,peyre2018computational} on GPUs. For the proposed method, we computed the tree-Wasserstein loss on the CPU and then passed it to the GPU to compute the gradient. Then, we set the number of iterations of the Sinkhorn-Knopp algorithm to 10 and the regularization parameter to $50$, respectively. 
\subsection{Synthetic data} \label{subsec-toy}
 We generated a synthetic dataset that comprises pairs of a real vector and a target probability distribution on the nodes of a randomly generated tree. This dataset was created as follows: First, we defined a parametric distribution on a graph. Given a graph, $G = (V, E)$, the shortest path metric, $d_G$, and the probability distribution, $F_{vu\sigma}$, over $V$ parameterized by $v, u \in V, \sigma > 0$ is defined as:
\begin{align*}
   F_{vu\sigma}(s) = \frac{1}{C}(\exp{\frac{d_G(v, s)}{\sigma^2}} + \exp{\frac{d_G(u, s)}{\sigma^2}}) \nonumber\\
    C = \sum_{s \in V}(\exp{\frac{d_G(v, s)}{\sigma^2}} + \exp{\frac{d_G(u, s)}{\sigma^2}}).
 \end{align*}
 Algorithm \ref{alg:syndata} shows the algorithm used to generate the dataset used in the experiments. In this experiment, we prepared datasets with the distribution on a random tree with $1000$ nodes using NetworkX \cite{SciPyProceedings_11}. The size of each of the training and testing datasets is $1000$. 
 We set the number of epochs to $500$ and the batch size to $10$, and we fixed the learning rate at $.001$. We reported the average scores of the experiments using $10$ different random seeds.
 
\vspace{.05in}
\noindent {\bf Predictive model:} We adopted the following model for class $\ell$:
 \begin{align*}
     \boldh_\boldtheta(\boldx)^{(\ell)} = \frac{\exp(\boldw_\ell^\top \boldx + b_\ell)}{\displaystyle \sum_{j}\exp(\boldw_j^\top \boldx + b_j)},
 \end{align*}
where $\mathbf{w}_i, b_i$ are learnable parameters.

 \vspace{.05in}
\noindent {\bf Evaluation Metric:} To evaluate predictions from various perspectives, we used the metric listed in Table \ref{tb:emdis}. Notably we adopted the {\bf exact} Wasserstein distance, called \textit{Wasserstein}, between the prediction and ground-truth label distributions to assess the extent to which the ground metric was considered in the prediction. In these experiments, we used the Python Optimal Transport (POT) library \cite{flamary2017pot} to calculate the exact Wasserstein distance, and the weights of all the edges were set to $1$. The other evaluation metrics are the same as those used in \cite{geng2016label}.

The scores of the experiment with synthetic data are presented in Table \ref{tb:syntree}. The proposed linear combinations of $\mathcal{KL}$ and $\mathcal{TW}$ outperformed the others in terms of \textit{Wasserstein} and \textit{Chebyshev} metric, but they performed poorly in terms of the other metrics.

\subsection{BlurbGenreCollectionEN} \label{subsec-blur}
In this study, we used the BlurbGenreCollectionEN\footnote{\scriptsize\url{https://www.inf.uni-hamburg.de/en/inst/ab/lt/resources/data/blurb-genre-collection.html}}\cite{cortes1995support, lewis2004rcv1} dataset for performing experiments with real data.
It comprises advertising descriptions of books from the Penguin Random House webpage. Each instance has one or multiple labels that are hierarchically structured. Because the hierarchical structure of these data is a {\it forest} and not a {\it tree}, we added a root node to the hierarchical tree. Of the total $91,892$ data samples $64$\%, $16$\% and $20$\% were used in the train, validation, and test sets, respectively. We set the number of epochs to $100$ and the batch size to $100$, and we fixed the learning rate to $.001$. We reported the average scores and standard deviations of the experiments using $10$ different random seeds.

\vspace{.05in}
\noindent {\bf Predictive model:} We adopted a {\it long-short-term-memory} (LSTM) \cite{hochreiter1997long} model with a hidden state size of $200$. Because LSTM can efficiently learn long-term dependencies of time-series data, it has often been used in the natural-language processing domain \cite{yin2017comparative, kuncoro2018lstms}. Additionally, we used {\it fastText} \cite{bojanowski2017enriching, joulin2017bag} for word embeddings. A fully connected layer exists before the output layer, and the output function is a softmax function.
 
\vspace{.05in}
\noindent {\bf Evaluation metric:} We evaluated prediction accuracy using three metrics, namely pseudo-recall, top-$k$ cost, and receiver operating characteristic area under the curve (ROC-AUC).
Pseudo-recall is defined as $\frac{|\calP \cup \calL|}{|\calL|}$, where $\calL$ denotes the set of ground-truth labels, and $\calP$ is a set that comprises $L = |\calL|$ labels in descending order of the probability score. 

Top-$k$ cost is defined as:
\begin{align*}
    \frac{1}{K}\sum_{k=1}^{K} \min_{\ell \in \calL} d(\ell_{p_k}, \ell),
\end{align*}
where $\ell_{p_k}$ denotes the label with the $k$-th highest probability score. This metric measures how close the predicted top-$k$ labels are to the ground-truth labels. We calculate ROC-AUC using the output distribution of each model as a score vector, which is assigned $1$ on the ground truth labels or $0$ on the other labels. \newpage\noindent
Table \ref{tb:blurb} presents the comparison results. Both regularization terms ($\mathcal{W}^1$ and $\mathcal{TW}$) did not have a significant impact on the results.

 \begin{table}[tb]
\small
\begin{center}
    \begin{tabular}{|c|c|} \hline
    \textit{Canberra} & $\sum_{\ell=1}^L \frac{|\boldh_\boldtheta (\boldx)^{(\ell)} - \boldp_{\boldy}^{(\ell)}|}{\boldh_\boldtheta (\boldx)^{(\ell)} + \boldp_{\boldy}^{(\ell)}}$ \\
    \textit{Chebyshev} & $\max_i |\boldh_\boldtheta (\boldx)^{(\ell)} - \boldp_{\boldy}^{(\ell)}|$  \\
    \textit{Clark} & $\sqrt{\sum_{\ell=1}^L\frac{(\boldh_\boldtheta (\boldx)^{(\ell)} - \boldp_{\boldy}^{(\ell)})^2}{(\boldh_\boldtheta (\boldx)^{(\ell)} + \boldp_{\boldy}^{(\ell)})^2}}$ \\
    \textit{Cosine} & $\frac{\sum_{\ell=1}^L \boldh_\boldtheta (\boldx)^{(\ell)} \boldp_{\boldy}^{(\ell)}}{\sqrt{\sum_{\ell=1}^L (\boldh_\boldtheta (\boldx)^{(\ell)})^2}\sqrt{\sum_{\ell=1}^L (\boldp_{\boldy}^{(\ell)})^2} }$\\
    \textit{Intersection} & $\sum_{\ell=1}^L \min(\boldh_\boldtheta(\boldx)^{(\ell)}, \boldp_{\boldy}^{(\ell)})$ \\
    \textit{Kullback-Leibler} & $\sum_{\ell=1}^L \boldp_{\boldy}^{(\ell)} \ln \frac{\boldp_{\boldy}^{(\ell)}}{\boldh_\boldtheta(\boldx)^{(\ell)}}$ \\\hline
    \end{tabular}
    \caption{Evaluation metrics for LDL. $\boldh_\boldtheta (\boldx)$ is the predicted distribution of $\boldx$, and $\boldp_\boldy$ is the ground truth distribution of a label $\boldy$.\label{tb:emdis}}
\end{center}
\end{table}

\begin{algorithm}[htb]
\caption{Generating a synthetic dataset \label{alg:syndata}}
\DontPrintSemicolon
  Generate a random tree : $G = (V, E)$, where $V = \{s_1,...,s_l\}$\\
  $W_1 \leftarrow (n \times m)$\text{-dim random matrix} \\
  $W_2 \leftarrow (m \times (l+1))$\text{-dim random matrix} \\
  \For{$i = 1$ to $N$}
    {
        $x_i \leftarrow n\text{-dimensional random vector}$ \\
        $x_i \leftarrow \frac{1}{1 + \exp(-W_1x_i)}$ \\
        $x_i \leftarrow \frac{1}{1 + \exp(-W_2x_i)}$ \\
        $\sigma \leftarrow 10\mathbf{x_i}^{(l + 1)}$\\
        $j \leftarrow \text{argmax}_{1 \leq j \leq l} x_i^{(j)}$; \ $v \leftarrow s_j$\\
        $k \leftarrow \text{argmin}_{1 \leq k \leq l} x_i^{(k)}$; \ $u \leftarrow s_k$\\
        $p_G(s) \leftarrow F_{vu\sigma}(s), \forall s \in V$
    }
    return $\{(x_i, p_G(V))\}_{i = 1}^N$
\end{algorithm}

\begin{table}[t]
    \centering
    \caption{Comparison of computational efficiency. \label{tb:comeff}}
	\scalebox{0.7}{
		\begin{tabular}{|c|c|c|c|} \hline
        $L$ & Loss & Time(s) & Memory \\ \hline \hline
        
        \multirow{3}{*}{$10^2$} & $\mathcal{TW}$ & \bf{0.0024} & \bf{1.58 MB} \\
        
        & $\mathcal{W}^1$ with GPU & 0.0062 & 3.32 MB \\
        & $\mathcal{W}^1$ with CPU & 0.0528 & 2.98 MB \\ \hline
        
        \multirow{3}{*}{$10^3$} & $\mathcal{TW}$ & 0.0126 & \bf{2.44 MB} \\
        
        & $\mathcal{W}^1$ with GPU & 0.0071 & 16.94 MB \\
        & $\mathcal{W}^1$ with CPU & 0.1279 & 7.08 MB \\ \hline
		
		\multirow{3}{*}{$10^4$} & $\mathcal{TW}$ & \bf{0.1204} & \bf{9.82 MB} \\
        
        & $\mathcal{W}^1$ with GPU & 0.5277 & 766.88 MB \\
        & $\mathcal{W}^1$ with CPU & 25.7985 & 1148.22 MB \\ \hline
		
		\multirow{3}{*}{$10^5$} & $\mathcal{TW}$ & \bf{1.6454} & \bf{66.00 MB} \\
        
        & $\mathcal{W}^1$ with GPU & - & (37.25 GB) \\
        & $\mathcal{W}^1$ with CPU & - & (40.00 GB) \\
        \hline
		\end{tabular}
	}
	\vspace{-.1in}
\end{table}

\subsection{Computational-efficiency comparison}
In the computational efficiency experiment, distributions with $10^2$, $10^3$, $10^4$, and $10^5$ supports were prepared. Subsequently, the computation time and memory required to calculate the loss of pairs of random probability distributions on the supports were measured. To avoid calculating a shortest-path distance matrix, we used the matrix ($\boldone\boldone^\top - \mathbb{I}$), where $\mathbb{I}$ denotes an identity matrix, as the distance matrix while computing the Wasserstein loss. Additionally, we used a random tree, with edge weights of $1$, as a tree metric while computing the tree-Wasserstein loss. We report the average scores of three measurements.

Table \ref{tb:comeff} presents the time and memory required to calculate the losses for various numbers of nodes. $\mathcal{TW}$ outperforms the other Wasserstein losses in terms of computation time and is significantly superior in terms of memory consumption. Although $\mathcal{W}^1$ that uses a GPU is faster than the others with $10^3$ supports, it cannot calculate the loss with $10^5$ supports because the required memory cannot be allocated.

\section{Conclusions}\label{sec-conclusion}
This study proposed the use of a tree-Wasserstein reguralizer for learning. The experimental results indicate that our proposed method can successfully predict the distributions of structured labels and that it outperforms existing Wasserstein loss calculation methods in terms of both computational speed and memory consumption.

\section*{Acknowledgments}

This work was supported by the JSPS KAKENHI Grant Number 20H04243 and 20H04244. This work was also supported by JST, ACT-X Grant Number JPMJAX200S, Japan.

\bibliography{arxiv}
\bibliographystyle{unsrt}

\clearpage

\end{document}